\documentclass{article}

\usepackage[final,nonatbib]{nips_2016}

\usepackage[utf8]{inputenc} 
\usepackage[T1]{fontenc}    
\usepackage{hyperref}       
\usepackage{url}            
\usepackage{graphicx}

\usepackage{wrapfig}
\usepackage{booktabs}       
\usepackage{amsfonts}       
\usepackage{nicefrac}       
\usepackage{microtype}      
\usepackage{amsmath}
\usepackage{physics}
\usepackage{booktabs}
\usepackage{tikz}
\usepackage{amssymb}
\usepackage{algorithm}
\usepackage{algpseudocode}
\usepackage[style=apa,backend=biber]{biblatex}

\usepackage{caption}
\usepackage[utf8]{inputenc}
\usepackage{pgfplots}
\DeclareUnicodeCharacter{2212}{−}
\usepgfplotslibrary{groupplots,dateplot}
\usetikzlibrary{patterns,shapes.arrows}
\pgfplotsset{compat=newest}
\tikzset{
    semithick/.style={line width=0.8pt},
}
\usepgfplotslibrary{groupplots}
\usepgfplotslibrary{dateplot}

\title{Plan of Thoughts: Heuristic-Guided Problem Solving with Large Language Models}

\author{
  Houjun Liu \\
  Department of Computer Science, Stanford University\\
    \texttt{houjun@stanford.edu}\\
} 

\begin{document}
\maketitle

\begin{abstract}
While language models (LMs) offer significant capability in zero-shot reasoning tasks across a wide range of domains, they do not perform satisfactorily in problems which requires multi-step reasoning. Previous approaches to mitigate this involves breaking a larger, multi-step task into sub-tasks and asking the language model to generate proposals (``thoughts'') for each sub-task and using exhaustive planning approaches such as DFS to compose a solution. In this work, we leverage this idea to introduce two new contributions: first, we formalize a planning-based approach to perform multi-step problem solving with LMs via Partially Observable Markov Decision Processes (POMDPs), with the LM's own reflections about the value of a state used as a search heuristic; second, leveraging the online POMDP solver POMCP, we demonstrate a superior success rate of $89.4\%$ on the Game of 24 task as compared to existing approaches while also offering better anytime performance characteristics than fixed tree-search which is used previously. Taken together, these contributions allow modern LMs to decompose and solve larger-scale reasoning tasks more effectively.
\end{abstract}

\section{Introduction}
\label{sec:orga238b96}
Recent advances of language models (LMs) introduced the possibility of in-context, few or zero-shot reasoning (\cite{brown_language_2020}) using LMs without much or any fine tuning.

Yet, classically, LM decoding takes place in a left-to-right fashion, auto-regressively resolving one token at a time by sampling from the output distribution of possible next words without multi-step planning. 

Work in LM agents have taken steps to solve more complex problems that would typically require multi-step reasoning even while using this direct decoding approach. The simplest idea, named "chain-of-thoughts" (CoT), involves forcing the LM at decode time to begin the decoding process with natural language reasoning about its actions (\cite{wei_chain--thought_nodate}). The method has contributed to the creation of powerful language agents (\cite{yao_react_2023}) that can reason about complex actions.

Despite the relative success of CoT, the scheme still does not support any kind of backtracking as it samples directly from the LM's posterior distribution. When a problem requires a significantly large number of steps to solve, issues relating to "de-generation" (\cite{holtzman_curious_2020}) becomes increasingly prevalent: whereby, naive maximizations of sequence likelihood results in a most likely sub-phrase being repeated which does not contribute to increased information or progress on a problem.

Recent theoretical work suggests these types of degeneration arises due to the distortion of output probability density caused by the last-layer softmax projection into the probability simplex (\cite{finlayson_closing_2023}): due the lower degrees of freedom offered by a probability syntax, both high and low tails of the latent next-word distribution becomes emphasized in the output probability distribution.

To address this, recent approaches such as Tree of Thoughts (ToT) (\cite{yao_tree_2023}) have separated the process of next-step proposal ("thinking") and choosing the actual step to take given a situation ("reasoning"). This separate allows the representation of a problem through only short decoding sequences that are less prone to degeneration, while allowing a separate LM call to score the value of being at any given partial solution through a maximum-likely single-token output that is less likely to be distorted.

In this work, we extend the ToT prompting scheme to formalize this process of "thinking" and "reasoning" via a language model as a Partially Observable Markov Decision Process (POMDP). We call this decoding scheme the Plan of Thoughts (PoT).

The key underlying assumption of the proposed PoT scheme involves the claim that LMs are able to make judgments about the value of a subsequence towards solving a problem by analyzing the likelihood of a particular sequence against a judgment of value. This assumption is supported by the existence of reinforcement learning formulations of LM-on-LM output verification---both for reasoning (\cite{verma_chai_2022}) and hallucination (\cite{liu_token-level_2022})--as well as the use of LM-inferred state value heuristics in the ToT approach.

We leverage this assumption by, similar to ToT, using the LM's evaluation of the likelihood of a sequence (similar to LM "scoring" of a "thought" in ToT) as a heuristic for the coherence and reasoning within a subsequence of LM output---forming a "self reflection" scheme similar to other LM-scoring schemes previously proposed (\cite{paul2023refiner,shinn2023reflexion}). Yet, differing from ToT, we explicitly formulate this scoring by an LM as an "observation" of an unobservable underlying latent understanding of the input sequence.

By solving the PoT problem with the anytime POMCP solver (\cite{silver2010monte}), we further demonstrate that PoT exhibits stronger anytime characteristics on the Game of 24 task as compared to ToT while maintaining performance that is comparable to ToT and superior to CoT. We were able to obtain these results at lower costs to ToT evaluations by using a hybrid language modeling approach by using a larger language model, GPT-4 (\cite{achiam2023gpt}), for posterior sampling and evaluation while using a smaller language model, GPT-3.5-Turbo-Instruct (\cite{brown_language_2020}), as the "thought" generator.

\section{Tree of Thoughts}
\label{sec:orgb74c944}

We provide here a short summary of the Tree of Thoughts (ToT) (\cite{yao_tree_2023}) approach which we improve upon in our work. ToT offers a scheme to enable multi-step reasoning with LMs by presenting a decomposition of multi-step LM reasoning into individual steps which is then combined through classic approaches in search and planning.

Specifically, ToT represents a given problem as a finite-horizon planning problem which it then solves in four broad steps. 

\textbf{Thought Decomposition}: by leveraging problem-specific characteristics, each problem is decomposed into distinct, incremental steps towards a solution. For the "Game of 24" task, for instance, each "thought" is considered a line of equation which contributes to the overall task of combining four numbers to reach 24.

Now, let \(p_{\theta}\) be our language model, \(s_{j}^{(i)}\) be thought candidate \(j\) of step \(i\) of a decomposed problem, \(s_{*}^{(i)}\) the optimal thought to continue from at step \(i\), \(\tau_{ *} = \qty[s^{(1)}_{ *}, ..., s^{(n)}_{ *}]\) a "solution" to a given problem. 

\textbf{Thought Generation}: multiple, initial short decodings of a LM---sampling from \(s' \sim p_{\theta}^{thought}\qty(s^{(i+1)} | s^{(i)})\) is obtained which forms a series of next states ("thoughts") which encodes a partial step towards the solution which is reachable at any given state. 

\textbf{Thought Evaluation}: another LM call rates each of the possible next states for their chance in reaching the solution; specifically, we ask the LM to reason about a given state by calculating the posterior probabilities of predicting a specific judgement of value (the words "sure"\emph{"likely"}"impossible") given that state; that is: \(V(s_{j}) = \arg\max_{o} \{p^{value}_\theta(o|s_{j}), o \in \{ sure, likely, impossible\}, \forall j \in 1 ... n\).

\textbf{Problem Solving}: finally, given this heuristic, solving a specific problem in ToT involves using a search-and-planning scheme (specifically, DFS or BFS) to cycle between generation and evaluation of thoughts until a terminal thought is reached. Branches on the DFS tree is pruned if they are voted as "impossible".

By combining explicit planning and LM reasoning, this approach achieved state-of-the-art results on the Game of 24 and other difficult natural-language tasks such as a crossword. However, the ToT approach does not incorporate any form of heuristic-guided preferential planning between different "possible" states---in contrast to dynamic approaches which preferentially explore sequences of high probability of success.

\section{Methods}
\label{sec:orgd687036}

\subsection{Problem Formulation}
\label{sec:orgff64743}
Our work formalizes and augments the stepwise decoding scheme proposed by ToT as a Partially Observable Markov Decision Process (POMDP) (\cite{kaelbling_planning_1998}). A POMDP is a search and planning formulation which emphasizes the uncertain nature of intermediate steps by formalizing each problem into a tuple \((\mathcal{S}, \mathcal{A}, \mathcal{O}, \mathcal{T}, \mathcal{R}, \gamma, s_0)\).

We specifically define our problem formulation as follows:

\begin{itemize}
\item \(\mathcal{S} = S \times U\), where \(s \in S\) is each sub-step of our decomposed problem and \(u \in U\) representing the unmeasurable, true underlying value being estimated by \(V(s)\) in ToT representing the usefulness of a particular thought
\item \(\mathcal{A} = [a_0, a_1, a_2]\), a discrete set of possible problem-solving actions---to "continue" expanding a particular branch \(a_0\), to "rollback" to previous branch \(a_1\), or to "think" a new thought at the current branch \(a_2\).
\item \(\mathcal{O} \in S \times U\), exactly the same as \(\mathcal{S}\), but instead of the unobservable underlying value of a given \(s\) we obtain \(V(s)\) instead from the language model by asking the language model for its judgement regarding a state; importantly, because the observations are simply a sample on the distribution, we can directly use \(V(s_{j}) \sim p^{value}_\theta(o|s_{j}), o \in \{ sure, likely, impossible\}, \forall j \in 1 ... n\)
\item \(\mathcal{T}\) is given deterministically given the state and action---"continue" appends the current state to the solution trajectory, yielding a new subproblem, and calculates a new thought; "rollback" pops the last item in the solution trajectory back into the current state and reverts to the previous subproblem; "think" reformulates a new state given the current subproblem
\item \(\mathcal{R}\) is given by a language model evaluator given a final trajectory, where: \(r_{\max}\) is given if the LM believes a trajectory successfully solves the problem, and \(r_{\min}\) is given if the LM believes a trajectory failed to solve the problem
\end{itemize}

Lastly, for this problem, we set discount \(\gamma\) to \(1\) to maximize joint reward, and \(s_0\) would be the initial, unsolved problem.

\subsection{Modified POMCP}
\label{sec:orgbc0ee0a}
To solve the formalization given above in an actual problem, we chose the POMCP solver (\cite{silver2010monte}). This solver is chosen for three primary reasons.

First, by only needing actions and observation sequences as input, the solver requires no explicit distribution on observation and transitions, meaning that simply making concrete samples from the language model posterior is enough to take advantage of its distributional nature.

Second, the POMCP solver has excellent anytime performance characteristics; the search tree for possible solutions will prioritize most possible solutions as rated by intermediate value, but can expand to (at the worst case) an exhaustive search of all possible intermediate states. In particular, easier problems will have stronger heuristic signals, which typically will take less time to solve; this means that a cutoff could be specified by the user to control the speed/accuracy trade-off in solving a given problems.

Specifically, a POMCP solver collects a tree based on sequences of actions and their resulting observations \(h = \{a_1, o_1, ...\}\). When planning for a specific action, the scheme samples a series of possible next states from a generative model given your current state and action \(s' \sim T(\cdot | s,a)\) and calculates reward \(R(s,a)\) from current state.

Once this procedure grows the tree to a certain depth, a point-wise value estimate is calculated from a roll-out.

For this specific problem, we modify the typical "rollout" rollout procedure by essentially performing CoT reasoning with a weighted average of the possible rewards in obtained in the end:

\begin{algorithm}
\caption{obtain a value estimate at some leaf state $s_{f}$}\label{alg:cap}
\begin{algorithmic}
\Ensure $d > f$
\State $s = s_{f}$
\State $L \gets d-n$ \Comment{Depth Remaning in Rollout}
\State $\tau \gets \{s_0, \dots, s_{f}\}$ 
\While{$L \neq 0$}
\State $\tau \gets \tau \cup \left\{\arg\max_{s'} \qty(p^{thought}_{\theta}(s'|s))\right\}$
\State $s = s'$
\State $L \gets L-1$ 
\EndWhile
\State $V = \frac{R_{\max} \cdot p_{\theta}^{evaluate}(\tau^{*}|\tau)+R_{\min} \cdot p_{\theta}^{evaluate}(\neg\tau^{*}|\tau)}{R_{\max}+R_{\min}}$\Comment{Posterior Weighted Average of Possible Reward}
\State \Return $V$
\end{algorithmic}
\end{algorithm}

where, \(p_{\theta}^{thought}\) is the "thought" generation prompt previously discussed, and \(p_{\theta}^{evaluate}\) is the evaluation prompt to check if a particular trajectory truly solves the target task which is also used in reward calculation. Recall that \(\tau^{*}\) represents a trajectory which answers the given question correctly.

As CoT is a reasonably performant reasoning procedure that is relatively lightweight to compute, we believe it would serve to \emph{raise} the lower bound of possible values, and therefore aid the speed of solution in POMCP.

\subsection{Task Setup}
\label{sec:orgfe23ed7}

\begin{figure}[H]
    \centering
\includegraphics[width=.9\linewidth]{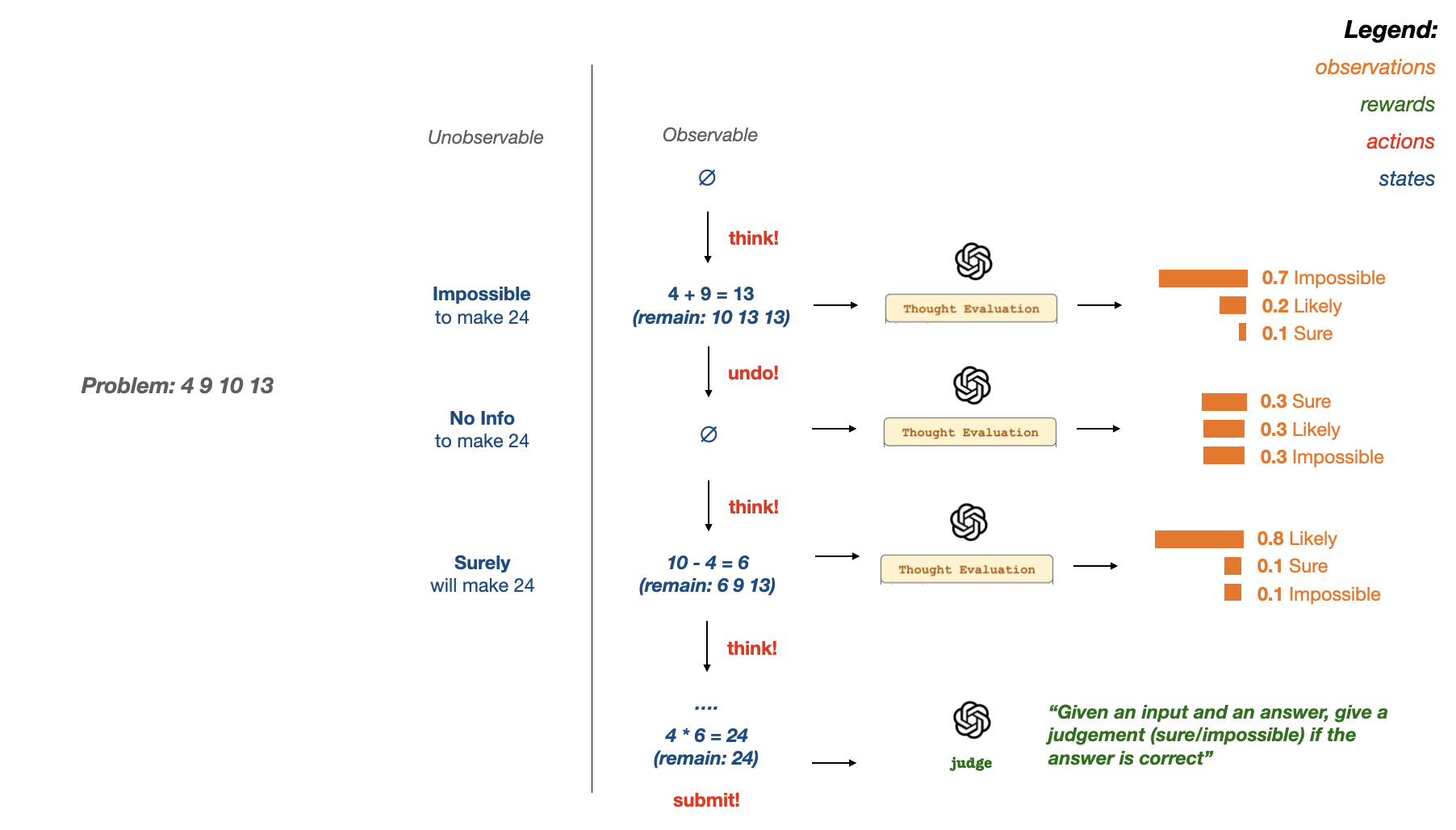}
\caption{Trajectory of PoT solver interacting with a ``Game of 24''}
\end{figure}

Similar to ToT, we are going to use the Game of 24 as a difficult multi-step reasoning task with which to test the scheme proposed.

The Game of 24 is a mathematical reasoning game, which uses four numbers and basic arithmetic operations to obtain a value of 24. For instance, for the problem of \texttt{4 9 10 13}, a solution trajectory in terms of subproblems used in our work may look as follows:

\begin{itemize}
\item \(s_0\): subproblem: 4 9 10 13
\item \(s_1\): \(13-9=4\), subproblem: 4 4 10
\item \(s_2\): \(10-6 = 6\), subproblem: 4 6
\item \(s_3\): \(4 \cdot 6\), subproblem: 24
\end{itemize}

which concludes the solution.

\textbf{Data Source}: in order to maintain comparability to ToT, we leverage the exact dataset curated by Yao et. al. scraped from 4nums.com as our problem set as well. Importantly, the data is sorted by rate of difficulty (as measured by weighted average time for solution)

\textbf{Benchmark}: the "success rate" metric reported involves the success rate across 100 games, corresponding to the metric reported by ToT. Additionally, we also report time-to-solve metrics as measured by the time between the initialization of an empty POMCP tree to obtaining a proposed solution from the scheme.

\textbf{Language Modeling}: distinct from ToT, to perform language model, we use \emph{two separate language models}. \(p_{\theta}^{evaluate}\) and \(p_{\theta}^{value}\) (for \(\mathcal{R}\) and \(\mathcal{O}\) respectively) were computed using GPT-4-Turbo (1106), and \(p_{\theta}^{thought}\) was computed using GPT-3.5-Turbo-Instruct (0914). This hybrid approach allows for single-token only inference on the larger GPT-4 models, affording dramatic performance improvements. 

\textbf{Solving}: we performed language model inference through the OpenAI Azure Cognitive Services API and used the POMDPs.jl, BasicPOMCP.jl Julia packages for the orchestration of the solver.

\section{Results}
\label{sec:org71768e9}
\vspace{-3em}
\begin{minipage}{0.3\textwidth} \centering
\begin{table}[H]
\centering
\vspace{2em}
\begin{tabular}{ll}
\toprule
\textbf{Method} & \textbf{Success} \\ \midrule
CoT & 4.0\%\\[0pt]
ToT (b=1) & 45\%\\[0pt]
ToT (b=5) & 74\%\\[0pt]
PoT (ours) & \textbf{89.4}\%\\[0pt]
\bottomrule
\end{tabular}
\vspace{2em}
\caption{Game of 24 Results; CoT and ToT results from \cite{yao_tree_2023}.}
\label{tab:results}
\end{table}
\end{minipage}
\hfill
\begin{minipage}{0.7\textwidth} 
\begin{figure}[H]
\centering
\resizebox{0.45\linewidth}{!}{\input{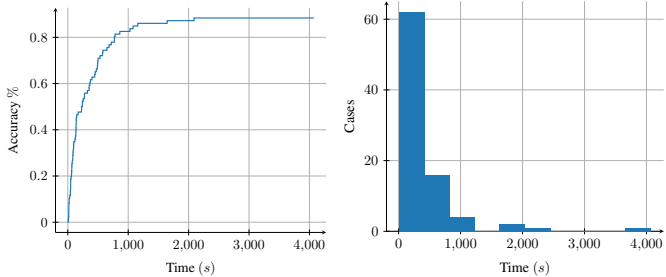}}
\resizebox{0.45\linewidth}{!}{
\begin{tikzpicture}

\definecolor{darkgray176}{RGB}{176,176,176}
\definecolor{steelblue31119180}{RGB}{31,119,180}

\begin{axis}[
tick align=outside,
tick pos=left,
x grid style={darkgray176},
xlabel={Time \(\displaystyle (s)\)},
xmajorgrids,
axis lines=left,
xmin=-195.748862302303, xmax=4277.72893885374,
xtick style={color=black},
y grid style={darkgray176},
ylabel={Cases},
ymajorgrids,
ymin=0, ymax=65.1,
ytick style={color=black}
]
\draw[draw=none,fill=steelblue31119180] (axis cs:7.59103775024417,0) rectangle (axis cs:414.270837855339,62);
\draw[draw=none,fill=steelblue31119180] (axis cs:414.270837855339,0) rectangle (axis cs:820.950637960434,16);
\draw[draw=none,fill=steelblue31119180] (axis cs:820.950637960434,0) rectangle (axis cs:1227.63043806553,4);
\draw[draw=none,fill=steelblue31119180] (axis cs:1227.63043806553,0) rectangle (axis cs:1634.31023817062,0);
\draw[draw=none,fill=steelblue31119180] (axis cs:1634.31023817062,0) rectangle (axis cs:2040.99003827572,2);
\draw[draw=none,fill=steelblue31119180] (axis cs:2040.99003827572,0) rectangle (axis cs:2447.66983838081,1);
\draw[draw=none,fill=steelblue31119180] (axis cs:2447.66983838081,0) rectangle (axis cs:2854.34963848591,0);
\draw[draw=none,fill=steelblue31119180] (axis cs:2854.34963848591,0) rectangle (axis cs:3261.029438591,0);
\draw[draw=none,fill=steelblue31119180] (axis cs:3261.029438591,0) rectangle (axis cs:3667.7092386961,0);
\draw[draw=none,fill=steelblue31119180] (axis cs:3667.7092386961,0) rectangle (axis cs:4074.38903880119,1);
\end{axis}

\end{tikzpicture}}
\caption{a) ROC curve by time b) Count by time to solve}
\label{fig:roc}
\end{figure}
\end{minipage}

Table \ref{tab:results} highlights the results we obtained by applying the PoT scheme on the Game of 24 problem. Specifically, we demonstrate a problem solving success rate which exceeds existing approaches---an in particular exceeds the results from CoT (\cite{wei_chain--thought_nodate}) by a significant margin.

Notably, because ToT used an exogenous number of search breadth, it is likely that with higher branches the approach can yield improved accuracy. Because POMCP has a dynamic branching characteristic to BFS---making a direct comparison difficult---we additionally report the relationship between \textit{time} spent in problem solving and accuracy in Figure \ref{fig:roc}. By setting the longest time of an hour ($3600$ seconds) as a threshold of $1$ and interpolating time linearly, the accuracy results we obtain scores an AUC value of $81.4\%$.

Importantly, Figure \ref{fig:roc}.b highlights that a significant amount (specifically, $83.7\%$) of the cases are solved within 10 minutes---representing only $20\%$ of the total time alloted to solve; this property demonstrates the anytime characteristic from PoT's use of POMCP.

\section{Conclusion}
\label{sec:org299bd16}

In this work, we present Plan of Thought (PoT), a novel language model decoding scheme that extends Tree of Thoughts (ToT) (\cite{yao_tree_2023}) which leverages a large language model's self-reflective reasoning capabilities (\cite{paul2023refiner,shinn2023reflexion}) to guide multi-hop reasoning about a problem.

We formalize our approach through the POMDP framework (\cite{kaelbling_planning_1998}), and demonstrate superior performance of our approach on the Game of 24 task compared to existing approaches using the online POMCP (\cite{silver2010monte}) solver. Because of the anytime behavior of POMCP, we are able to solving a large portion of the test problems in significantly less time than the maximum alloted. 

We were able to demonstrate the results presented using a hybrid language modeling approach by using GPT-3.5-Turbo-Instruct as our state generator and GPT-4-Turbo for evaluation only: unlike previous approaches which only used a single, fixed language model (usually GPT-4). Taken together, these properties makes PoT a more flexible approach to solving basic multi-step reasoning tasks as compared to previous approaches---allowing for contemporary LMs to solve more complex problems. 

Notably, as the previous ToT approach requires seaching on a tree of a fixed-breadth, it is likely that the approach will converge to similar performance as the PoT scheme presented here at higher choices of $b$. Importantly, however, the anytime property of the approach presents here indicates that PoT can scale its solution time dynamically to the number of ``thoughts'' required for each problem as shown by the AUC value of $81.4\%$ of our approach using time as a threshold, while ToT will take a fix-time approach to generate and evaluate an entire tree regardless of solution complexity. 

In its current form, our work has two key limitations. First, similar to that proposed by ToT, the approach is still \emph{significantly} more computationally expensive than CoT or other direct decoding approaches; therefore, these search techniques is likely unnecessary for problems which can be solved with high accuracy using simpler techniques. Second, posterior distributions---even when taking only top-k samples for extremely small k---are still meaningful only on billion-parameter models if used without additional fine tuning (\cite{hu2023prompting}): making the heuristic-driven performance improvements of PoT limited in scope. With additional fine-tuning of surrogate value models, PoT could likely perform dramatically more efficiently while obtaining its positive characteristics in solution quality.

\printbibliography

\end{document}